\documentclass[3p, 11pt]{elsarticle}
\usepackage{lineno}
\usepackage[english]{babel}
\usepackage{natbib}
\usepackage{amssymb}
\usepackage{setspace}
\usepackage{lipsum, color}
\usepackage{amsmath}
\usepackage{natbib}
\usepackage{multirow}
\usepackage{longtable}
\usepackage{float}
\usepackage{adjustbox}
\usepackage{cuted}
\usepackage{float}
\usepackage{chngcntr}
\usepackage{mathptmx}
\usepackage{setspace}
\usepackage{isomath}
\usepackage{subfig}
\usepackage{textcomp}
\usepackage{bm}
\usepackage{lineno}

\numberwithin{equation}{section}

\biboptions{authoryear}

\makeatletter
\def\ps@pprintTitle{%
   \let\@oddhead\@empty
   \let\@evenhead\@empty
   \let\@oddfoot\@empty
   \let\@evenfoot\@oddfoot
}
\makeatother

\begin{document}

\begin{frontmatter}

\title{Identification of Choquet capacity in multicriteria sorting problems through stochastic inverse analysis}


\author[rvt]{Renata~Pelissari\corref{cor1}}
\ead{renatapelissari@unicamp.br}

\author[rvt]{Leonardo~Tomazeli~Duarte}
\ead{leonardo.duarte@fca.unicamp.br}

\cortext[cor1]{Corresponding author}

\address[rvt]{School of Applied Sciences, UNICAMP, Pedro Zaccaria 1300, 13484-350 Limeira, SP, Brazil}

\begin{abstract}
In multicriteria decision aiding (MCDA), the Choquet integral has been used as an aggregation operator to deal with the case of interacting decision criteria. While the application of the Choquet integral for ranking problems have been receiving most of the attention, this paper rather focuses on multicriteria sorting problems (MCSP). In the Choquet integral context, a practical problem that arises is related to the elicitation of parameters known as the Choquet capacities. We address the problem of Choquet capacity identification for MCSP by applying the Stochastic Acceptability Multicriteri Analysis (SMAA), proposing the SMAA-S-Choquet method. The proposed method is also able to model uncertain data that may be present in both decision matrix and limiting profiles, the latter a parameter associated with the sorting problematic. We also introduce two new descriptive measures in order to conduct reverse analysis regarding the capacities: the Scenario Acceptability Index and the Scenario Central Capacity vector. 
\end{abstract}

\begin{keyword}
Multicriteria sorting problem, Preference elicitation, Capacity identification, Choquet integral, Stochastic Multicriteria Acceptability Analysis
\end{keyword}
\end{frontmatter}

\onehalfspacing

\section{Introduction}

    \noindent This paper considers a multicriteria sorting problem (MCSP) with $m$ alternatives $A=\{a_1, \ldots, a_m\}$ evaluated in terms of $n$ criteria, $g_1, \ldots, g_n$, to be sorted into $k$ predefined and ordered categories, $K_1, \ldots, K_k$. When the considered problem has interacting criteria, a widely used aggregation operator is the Choquet integral \citep{Marichal2000, GrabischLabreuche2010}. Analogously to most of multicriteria decision aiding (MCDA) methods that require a criteria weight vector from the decision-maker (DM), methods whose aggregation procedure is based on the Choquet integral require the prior identification of a capacity. Choquet capacities can be seen as weights to all coalitions of criteria and, therefore, differently from MCDA methods where $n-1$ weights are required, in the application of the Choquet integral $2^{n} - 2$ parameters are needed \citep{Marichal_Roubens2000}. 
    
    With the aim of reducing the number of parameters, the concept of 2-additive capacity was introduced in \cite{Grabisch1997}. The 2-additive property implies that there may be interaction between any two criteria but not among three or more and it enables to reduce the number of parameters from $2^{n} - 2$ to $n(n+1)/2$, which represents a significant reduction as $n$ grows. Therefore, a 2-additive capacity offers a good trade-off between ability of modeling interaction and complexity of the model. In this paper, we focus on the 2-additive model. 
    
    Even with the reduction of the numbers of parameters, the DM may still find difficult to define exact values for the capacities. Several works in the literature address this problem known as capacity elicitation or capacity identification by applying supervised, semi-supervised and unsupervised methods, as discussed in depth by \cite{Leo2018}. Supervised methods, also called preference disaggregation, or indirect elicitation in the MCDA field, require examples of decisions in order to estimate the capacities \citep{Angilella2010, Grabisch2008-2}. Although indirect methods are generally attractive for the DM, these exhibit a significant drawback when there is a relatively small reference set of examples \citep{GrabischLabreuche2010, DOUMPOS2011203}.
    
    In semi-supervised methods, examples of decision are not required but other type of preference information is needed such as relative importance of criteria and an indication that there is an interaction between two criteria. Examples of semi-supervised methods include works based on diamond pairwise comparison  \cite{Wu2015} and on the Stochastic Multicriteria Acceptability Analysis (SMAA) \cite{Angilella2015a}. 
    
    In unsupervised methods, the elicitation of the capacities is conducted without any prior information from the DM. The only information needed by the method is the decision matrix and the method is strongly based on statistical tools since the interactions between the criteria must be estimated from the observed data \citep{Leo2018}.
    
    As one can note in the literature, while the proposal of approaches for capacity identification has been focusing on ranking MCDA problems (where the objective is to obtain a ranking of the alternatives), very few similar approaches were proposed for MCSP. In the MCSP context, \cite{BENABBOU2017152} proposed an incremental capacity elicitation approach through a sequence of preference queries.
    
    Our aim is to propose a semi-supervised method for capacity identification in the MCSP context by applying SMAA. Since SMAA methodology is based on an inverse analysis of the space of feasible parameter values, we propose to apply SMAA in order to describe the capacities that make an alternative to be placed into a determined category. To make a reverse analysis possible, the SMAA methodology gives as result descriptive measures which are analyzed by the DM in order to obtain a final decision. Although many descriptive measures have been already proposed for different SMAA variants (see \cite{Pelissari2019}), there are still some limitations with regard to descriptive measures in SMAA sorting methods. 
    
    Firstly, the acceptability category index, which informs the probability of an alternative to be assigned to a given category, is the only descriptive measure provided by SMAA sorting methods \citep{Tervonen2009a, Lahdelma2010, PELISSARI2019235}. Based on this measure,  alternatives are usually assigned to the category with the highest acceptability category index and it is possible that the classification of two alternatives being resulting from different capacities which may be considered a drawback of SMAA sorting methods. Indeed, it is natural to assume that DMs expect that all alternatives being classified based on the same preference system. In order to overcome this limitation, our recommendation is then to analyze the feasible scenarios formed by the possible combinations of alternatives classification instead of analyzing the individual classification of each alternative. For that, we propose the scenario acceptability index which indicates the probability of occurrence of each feasible scenario.
    
    The second limitation regards the lack of a descriptive measure allowing the reverse analysis. While most of SMAA ranking methods calculates the central weights vector, a descriptive measure that shows the criteria weights which make an alternative ranks first, there is no similar measure for SMAA-sorting methods. We then present the Scenario Central Capacity vector which allows the reverse analysis identifying the capacity vector that makes a scenario possible.
    
    
    Therefore, our contribution is threefold: (1) proposal of SMAA-S-Choquet for capacity identification in multicriteria sorting problems; (2) proposal of the scenario acceptability index descriptive measure, introducing the concept of analysing scenarios; (3) proposal of the scenario central capacity vector for reverse analysis in SMAA-S-Choquet. It is important to highlight that both proposed descriptive measures can be easily modify to be used in any SMAA sorting methods. It is also worth mentioning that SMAA-S-Choquet also allows the application of the Choquet integral preference model when the decision matrix and other model parameters are defined by random variables instead of necessarily deterministic values.
    
    This paper is organized as follows. In Section 2, we introduce the Choquet integral preference model for multicriteria sorting problems. In Section 3, we propose the SMAA-S-Choquet method and the new descriptive measures. In Section 4 we conduct some numerical tests and present a numerical example. Conclusions and future research trends are presented in Section 5.
    
\section{Sorting with the Choquet integral preference model}

        \noindent Choquet integral is based on the concept of capacity (fuzzy measure). However, as presented by \cite{Grabisch2000}, it is also possible to express the Choquet integral in terms of other measures such as the generalized interaction index. For the 2-additive capacities case, the interaction index between two criteria $\{g_j, g_s\}$ is represented by $I_{js}$ and, when the generalized interaction index is calculated for a single criterion, say $g_j$, it gives rise to the Shapley importance index which is denoted by $I_j$ = $I(g_j)$. Still in the 2-additive capacities context, capacities are completely determined by the Shapley indexes $I_j$ and the interaction indexes $I_{js}$. 
        
        In this study, the Choquet integral is parametrized in the generalized interaction index domain. Such a representation admits a direct interpretation, as the Shapley has more practical interpretation. Indeed, the Shapley index $I_j$ quantifies the average importance of a single criterion $g_j$ for the decision process, similar to criteria weights in traditional MCDA methods. Moreover, $I_{ij}>0$ models a synergy between $g_j$ and $g_s$, $I_{js}<0$ represents a redundancy between $g_j$ and $g_s$, and $I_{js} = 0$ means that criteria $g_j$ and $g_s$ do not interact. Interaction indices have to meet the boundary and monotonicity coinstraints defined respectively by: 
        \begin{equation}
            \sum_{j=1}^{n} I_j = 1 \hbox{ and } I_{j} - \frac{1}{2}\sum_{j\neq s}|I_{js}| \geq 0, \forall j.
        \end{equation}
   
        \noindent The 2-additive Choquet integral preference model based on the generalized interaction index is expressed by
            \begin{eqnarray}
                CI(a) &=& \sum_{I_{js} > 0} \min \{g_j(a), g_s(a)\}I_{js} + \sum_{I_{js} < 0} \max \{g_j(a), g_s(a)\}|I_{js}| \nonumber\\&+& \sum_{j=1}^{n} g_{j}(a) (I_{j} - \frac{1}{2}\sum_{j\neq s}|I_{js}|).\label{choquet_shapley}
            \end{eqnarray}
            
        \noindent For the sorting problematic, $CI(a)$ has to be compared to reference profiles which characterize the predefined ordered categories. Reference profiles can be of two types: limiting profiles or central reference profiles. In this paper, we focus on the former, where categories are defined by two limiting profiles, the upper and lower limits of the category. A sorting procedure based on limiting profiles is also known as threshold-based driven sorting procedure. 
        
        Let $R = \{r_1, \ldots, r_{k+1}\}$ be the set of limiting profiles, in which $r_{1}$ and $r_{k+1}$ are the best and the worst, respectively. Since categories are completely ordered, each reference profile is preferred to the successive ones, i.e., $r_1 \succ r_2 \succ \ldots \succ r_k \succ r_{k+1}$. Moreover, performance of all alternatives of $A$ shall be between the worst and best limiting profiles. Therefore, we have formally that $\forall a_i \in A, \forall g_j \in G: g_j(r_1)\geq g_{j}(a_i) \geq g_{j}(r_{k+1})$. 
        
        The Choquet integral sorting formulation according to the threshold-based driven sorting procedure is then defined as follows \citep{BENABBOU2017152,GRECO20101455}:
        \begin{equation}\label{Choquet_sorting_procedure}
            a \in A \hbox{ is sorted into category } K_h, \hbox{ iff } CI(a) \in [r_{h-1}, r_{h}[. 
        \end{equation}
       
\section{The proposed method: SMAA-S-Choquet}

    We present the proposed method SMAA-S-Choquet starting by discussing its required input data. The decision matrix and the set of limiting profiles need to be completely defined, although they can be defined by uncertain values. For those situations, evaluations of the alternatives are defined by random variables $\xi_{ij}$ with a probability density function $f_{\xi}(X)$ in the space $X$: 
    $$X = \{X \in \mathbb{R}^m \times \mathbb{R}^n: \xi_{ij}, i=1, \ldots, m, j=1, \ldots, n\}.$$
    Uncertain limiting profiles are also represented by random variables $\psi$ with a probability density function $f_{Y}(\psi)$ in the space $Y$: $$Y = \{\psi \subseteq \mathbb{R}^{(k+1)} \times \mathbb{R}^n: \psi_{hj} - \psi_{lj} \leq 0, \forall h<l, h=1, \ldots, k, j=1, \ldots, n\}.$$
    
    Choquet capacities are not directly required. Indeed, our aim is the identification of the capacities. However, some preference information are still required from the DM. We shall suppose that the DM is able to provide some indirect preference information in terms of the interaction index, indicating which pairs of criteria are synergistic or redundant, and in terms of the Shapley index, indicating the relative importance of the criteria. This preference information then results in the following restrictions:
         \begin{itemize}
             \item Restrictions related to the DM preference information regarding interaction index:
                $$E^{DMP-I} 
                \begin{cases}
                 I_{js} > 0, \hbox{if criteria $j$ and $s$ are synergy,}\\
                 I_{js} < 0, \hbox{if criteria $j$ and $s$ are redundant.}                    \end{cases}
                $$
            \item Restrictions related to the DM preference information regarding the Shapley index:
                $$E^{DMP-S}
                \begin{cases}
                    I_j \geq I_s \hbox{, if } j \succeq s, \\
                    I_j > I_s \hbox{, if } j \succ s, \\
                    I_j = I_s \hbox{, if } j \sim s.
                \end{cases}
                $$
            \item In addition to the restrictions related to the DM preferences, restrictions that ensure monotonicity and boundary conditions shall also be considered for the capacity identification ($E^{MB}$):
                \begin{equation}
                    \sum_{j=1}^{n} I_j = 1 \hbox{ and } I_j - \frac{1}{2}|\sum_{j \neq s}| \geq 0, \forall j.\\
                \end{equation}
            \end{itemize}  
    \noindent We shall call compatible model a capacity whose interaction index representation satisfies the set of constraints $E = E^{DMP-I} \cup E^{DMP-S} \cup E^{MB}$. Total lack of preference information is possible only regarding the Shapley index. In this case, $E$ reduces to $E = E^{DMP-I} \cup  E^{MB}$.
    
    In the rest of this section, we describe the simulation scheme, presented in Algorithm 1, on which the SMAA-S-Choquet is based, and present the new descriptive measures proposed here. 
    
    The simulation processes starts starts by drawing criteria evaluations and limiting profiles from their probability distributions. Criteria evaluations $g_j(a_i)$ have to be sampled in order to be between the worst and best sampled limiting profiles. 

\begin{table}[htb]
    \caption{Algorithm 1: SMAA-S-Choquet simulation.}\label{tab:my_label}
    \begin{tabular}{|l|}\hline
       \textbf{Assume} Choquet integral sorting method as the decision model $M(\vec{a}, \vec{r}, \vec{I})$\\
        \textbf{repeat} \\
             ~~~\textbf{Draw} $\vec{\xi}$, $\vec{\psi}$ from their distributions; \\
             ~~~~~~~~~~~   $\vec{I}$ using HAR\\
             ~~~\textbf{Classify} the alternatives using $M(\vec{a}, \vec{r}, \vec{I})$\\
             ~~~ \textbf{Update} statistics about alternatives\\
        \textbf{until} Repeated $N$ times\\
        \textbf{Compute results} based on the collected statistics\\\hline
    \end{tabular}
\end{table}

    \noindent For capacity elicitation we apply a simulation process based on the Hit-And-Run (HAR) method \citep{Tervonen2013}. The Hit-And-Run sampling begins with the choice of one point inside the constrained capacity space $E$. At each iteration, a random direction is sampled that, with the considered position, generates a line. The intersection of this line with the boundary of space $E$ generates a line segment from which the next point is drawn uniformly. The HAR sampler is implemented using the hitandrun package for the R statistical software proposed by Tervonen and available at http://cran.r-project.org/web/packages/hitandrun/.
    
    Then, using the sampled values and the capacity vector elicited, the Choquet integral sorting formulation is applied. The obtained result is the classification of each alternative to the predefined categories. A ``scenario'' is here understood as the conjoined categorization of all alternatives. Denoting by $\vec{v}_t$ each possible scenario, we have the following definition: 
    $$\vec{v}_t= \{(K^{t}(a_1), K^{t}(a_2), \ldots, K^{t}(a_m))\}, $$
    \noindent where $K^{t}(a_i) = K_h$ is the category $h$ assigned to alternative $a_i$ in scenario $\vec{v}_t$. For simplification and for better visualization, each element of the vector scenario can also be denoted simply by the number of the category $K^{t}(a_i) = h$. Based on the scenarios, the following statistics are collected:
    \begin{itemize}
        \item $B_{t}:$ the number of times that scenario $\vec{v}_t$ occurs.
        \item $\vec{I}_{t}^{x}:$ Shapley index associate to scenario $t$, for each time $x=1,\ldots, B_{t}$ that scenario $\vec{v}_t$ occurs. 
    \end{itemize} 
    
    This process should be repeated a sufficient number of times in order to extract robust statistical information -- in previous works, 10,000 iterations were conducted \citep{Tervonen2007a}. After the last iteration of the simulation process, descriptive measures are computed in order to help the DM achieve a final decision. 
    
    The category acceptability index ($C_{i}^h, i=1,\ldots, m, h=1, \ldots, k$) is the measure given by SMAA sorting methods \citep{Tervonen2009a, Lahdelma2010, PELISSARI2019235}. It represents the probability of an alternative to be sorted into a category; very often, the alternatives are attributed to the category with the highest category acceptability value. For instance, let us consider a decision-making problem with two alternatives and two categories in which $C_{1}^1 = 0.9$ and $C_{1}^2 = 0.1$, and $C_{2}^1 = 0.38$ and $C_{2}^2 = 0.62$. Therefore, the DM may conclude that $a_1$ and $a_2$ should be sorted into categories $K_1$ and $K_2$, respectively. It is reasonable to assume that the DM wants to sort the set of all alternatives based on the same capacity vector, i.e, whether alternative $a_1$ is categorized to category $K_2$ considering that criteria 1 is more important than criteria 2, for instance, one expects that the categorization of the other alternative is also based on the fact that criteria 1 is more important than criteria 2. However, there is no guarantee that this occurs when using the category acceptability index. 
    
    In order to overcome this drawback, we propose the Scenario Acceptability Index (SAI), which is based on the category acceptability index, but it uses the concept of scenario (introduced right above) instead of individual classifications of the alternatives. While the category acceptability index represents the probability of an alternative to be sorted into a category, SAI represents the probability of a scenario occurring. Therefore, based on the collected statistics, estimates for the SAI are computed by
    $$SAI_t = \frac{B_{t}}{N},$$
    \noindent where $N$ is the total number of executions.
    
    The SAI of an impossible scenario is equal to 0, the SAI of any feasible scenario is between 0 and 1 and the sum of all SAI is equal to 1. The greater the value of $SAI_t$, the greater the probability of a scenario $\vec{v}_{t}$ occurring. Therefore, SAI is suggested to be used to support the prioritization of scenarios and, consequently, the indication of categories assigned to each alternative.  
    
    The capacity vector that makes a scenario possible is also a relevant criterion to be used by the DM for prioritizing scenarios. For that, we propose the Scenario Central Capacity vector which is based on the central capacity vector (introduced by \cite{Angilella2015}). While the latter represents the typical preferences that make an alternative preferred in a ranking problem, scenario central capacity vector $\vec{I}^{c}_{t}$ represents the typical DM preferences that make possible the occurrence of scenario $\vec{v}_{t}$. It is regarded as the expected centroid of the favorable capacity space and is defined:
    $$\vec{I}^{c}_{t} = \frac{\sum_{x=1}^{B_t}\vec{I}_{t}^{x}}{B_t}.$$
    \noindent The scenario central capacity vector allows a reverse analysis of capacities in the application of SMAA-S-Choquet. The proposed method SMAA-S-Choquet was implemented by the authors in the R statistical software. 
    
\section{Example and Experiments}

    \noindent The proposed method is illustrated by the following synthetic example inspired by \cite{Grabisch2005} and \cite{Angilella2015}. The dean of a technical school wants to categorized five students into 3 categories, accepted ($K_1$), waitlisted ($K_2$) and rejected ($K_3$), based on their grades on Mathematics and Statistics. A student good in Mathematics and in Statistics is of course well appreciated, but since students good in Mathematics are in general also good in Statistics, the dean does not want to overvalue students having good grades in both. Therefore, we can say that there is a negative interaction (redundancy) between Mathematics and Statistics.
    
    In order to apply the SMAA-S-Choquet, indirect preference information in terms of the interaction index is required. To represent the redundancy between Mathematics and Statistics the dean defined $I_{12} <0$. Defining preference information regarding Shapely index is not mandatory but it can be included in the model if the DM wishes, and the dean defines that the importance of both subjects are the same ($I_1 = I_2$).
    
    Students were evaluated based on 10 tests applied in each subject, as presented in Table \ref{tab:exemple} (Statistics grades were omitted for lack of space). Instead of considering the average grade, we assume the grades of each student in each subject as a random variable following a normal distribution whose parameters, mean and standard deviation, are estimated from the sample of 10 tests (Table \ref{tab:exemple}). When defining the limiting profiles, the dean demonstrated imprecision in relation to some of them, which was represented by interval values. The dean also assumes that the limiting profiles are the same for both subjects $R=\{10, [7.5, 8], [4.5, 5], 0\}$. 
    
\begin{table}[]
    \centering
    \caption{Students grades on Mathematics and Statistics.}
    \label{tab:exemple}
    \begin{tabular}{|c|cccccccccc|cc|cc|}\hline
          & \multicolumn{12}{c}{Mathematics - 10 tests} & \multicolumn{2}{|c|}{Statistics}\\\hline
        Students~&~$T_{1~}$ ~&~$T_{2~}$ ~&~$T_{3~}$ ~&~$T_{4~}$ ~&~$T_{5~}$ ~&~$T_{6~}$ ~&~$T_{7~}$ ~&~$T_{8~}$ ~&~$T_{9~}$ ~&~$T_{10}$ ~& Mean & sd. & Mean & sd.\\\hline
        $a_1$~&~6.7 ~&~8   ~&~6.5 ~&~6.6 ~&~3.9 ~&~4.3 ~&~1.2 ~&~3.8 ~&~5.5 ~&~5.4 ~&5.2 &1.9 &4.2 &0.8\\
        $a_2$ ~&~8.3 ~&~9.1 ~&~9.1 ~&~8.0 ~&~8.3 ~&~9.3 ~&~9.3 ~&~9.4 ~&~9.1 ~&~9.8 ~&8.9 &0.6 &9.3 &0.2\\
        $a_3$ ~&~3.1 ~&~4.5 ~&~4.0 ~&~4.4 ~&~4.5 ~&~5.2 ~&~3.8 ~&~4.0 ~&~3.7 ~&~4.3 ~&4.1 &0.6 &6.3 &0.7\\
        $a_4$ ~&~6.9 ~&~8.5 ~&~6.1 ~&~9.5 ~&~9.5 ~&~9.5 ~&~4.3 ~&~9.6 ~&~9.4 ~&~7.7 ~&8.2 &1.4 &6.9 &0.6\\
        $a_5$ ~&~9.1 ~&~6.8 ~&~6.3 ~&~9.8 ~&~8.2 ~&~9.8 ~&~8.4 ~&~8.2 ~&~8.3 ~&~7.9 ~&~8.3 ~&1.1 &4.7 &0.8\\\hline
    \end{tabular}
\end{table}    
Applying SMAA-S-Choquet, $t=40$ feasible scenarios were obtained. In Table \ref{tab:res_example}, we present the five main scenarios and their respective $SAI(t)$ and $\vec{I}^{c}_{t}$. 
    \begin{table}[]
        \caption{Descriptive measures of SMAA-S-Choquet for the five main scenarios and the category acceptability index for the five students.}
        \label{tab:res_example}
        \centering
        \begin{tabular}{|c|cc|c|ccc|}\hline
        & \multicolumn{2}{c|}{Descriptive measures} & \multicolumn{4}{|c|}{Category accceptability index}\\\hline
        \multicolumn{1}{|c|}{Scenarios} &  $SAI(t)$ (\%) & $\vec{I}^{c}_{t}$ & Students & $C_1^{i}$ & $C_2^{i}$ & $C_3^{i}$ \\\hline
        $\vec{v}_{1}=(2,1,2,1,2)$  & 18\% & (0.5, 0.5, -0.46) &$a_1$& 3&55&42\\
        $\vec{v}_{2}=(2,1,2,2,2)$  & 16\% & (0.5, 0.5, -0.39)&$a_2$ & 100&0&0\\
        $\vec{v}_{3}=(3,1,2,2,2)$  & 13\% & (0.5, 0.5, -0.34)&$a_3$ & 0&94&6\\
        $\vec{v}_{4}=(3,1,2,1,2)$  & 13\% & (0.5, 0.5, -0.42)&$a_4$ & 54&46&0\\
        $\vec{v}_{5}=(2,1,2,1,1)$  & 11\% & (0.5, 0.5, -0.62)&$a_5$ &34&66&0\\\hline
        \end{tabular}
    \end{table}
    Analyzing the scenario acceptability index $SAI$, the dean may choose either scenario $\vec{v}_{1}$ or $\vec{v}_{2}$ since both have the highest $SAI$ values ($SAI(1)=18\%$ and $SAI(2)=16\%$). To decide between these two, it is important to identify the scenario which capacity better describes the dean' preferences. Since the dean defined $I_1=I_2$, due to the monotonicity condition, the HAR approach generates $I_1=I_2=0.5$ for all scenarios and iteration, which leads the dean to focus on analyzing the interaction index $I_{12}$. One may realize that the higher the interaction index absolute value $I_{12}$, the more the alternatives are sorted into better categories. Indeed, the higher the absolute value $I_{12}$, the more redundant are considered the two subjects, and therefore, the more sufficient is a good grade in only one subject. Therefore, the dean should choose scenario $\vec{v}_{1}$ over $\vec{v}_{2}$ only if she wants to consider a higher interaction index absolute value.
    
    We also present the category acceptability index in Table \ref{tab:res_example} in order to compare it to the result obtained by using $SAI$. Based on the category acceptability index, the final categorization of alternatives would be identical to scenario $\vec{v}_{1}$. However, with this example, it becomes clear that this information (the category acceptability index) is not enough to support the DM since information regarding the capacities is also needed. 
    
    In order to verify the robustness of SMAA-S-Choquet, we performed a numerical experiment regarding sampling variability, on synthetic data. Specifically, we run each proposed model with 200 iterations, and the mean and standard deviation of scenario acceptability indices and scenario capacity vectors are computed. 
    
    The MCSP considered in this experiment has the following characteristics: two criteria, ten alternatives and three categories were considered; the same limiting profiles are defined by both criteria $R=\{10,7,5,3,0\}$; criteria 1 is less important than criteria 2, i.e, $I_1<I_2$; there is a redundancy between $g_1$ and $g_2$, i.e, $I_{12} < 0$. The decision matrix is randomly generated according to a uniform distribution in the range [0,10] (Table \ref{tab:evaluation_Numerical_Experiment}).
    \begin{table}[htb!]
    \centering
    \caption{Evaluation of alternatives used in the numerical experiment - randomly generated according to a uniform distribution in the range [0,10].}
    \label{tab:evaluation_Numerical_Experiment}
    \begin{tabular}{|l|cccccccccc|}\hline
     Criteria & $a_1$ & $a_2$& $a_3$& $a_4$& $a_5$& $a_6$& $a_7$& $a_8$& $a_9$& $a_{10}$\\\hline
     Mathematics ($g_1)$&~9.90~&~0.03~&~2.04~&~6.73~&~7.66~&~0.65~&~1.48~&~0.35~&~7.46~&~3.63 \\
     Statistic ($g_2$) &~~9.44 ~&~0.25~&~9.94~&~2.11~&~4.28~&~3.69~&~9.45~&~2.19~&~6.69~&~9.74\\\hline
    \end{tabular}
\end{table}

\begin{table}[h!]
    \centering
    \caption{Feasible categorization scenarios and their scenarios acceptability indices}
    \label{tab:categorization}
    \begin{tabular}{|c|c|c|c|c|}\hline
         & $SAI(t)$ (\%)& \multicolumn{3}{c|}{$\vec{I}^{c}_{t}$} \\\hline
         Scenario ($\vec{v}_{t}$)              & Mean$\pm$ sd & $I_1$& $I_2$& $I_{12}$ \\\hline
      $\vec{v}_{1}=(1,4,1,2,1,3,1,4,1,1)$  & 3.63 $\pm$ 0.32  &0.47$\pm$ 0.03 & 0.53$\pm$0.04 & -0.79$\pm$0.09\\
      $\vec{v}_{2}=(1,4,1,2,2,3,1,4,1,1)$  & 11.69 $\pm$ 0.89  & 0.41   $\pm$ 0.04 & 0.58$\pm$ 0.04 & -0.59$\pm$ 0.09\\
      $\vec{v}_{3}=(1,4,1,2,2,4,1,4,1,1)$  & 3.47 $\pm$ 0.30& 0.47 $\pm$0.03& 0.53$\pm$0.03 & -0.41$\pm$0.05\\
      $\vec{v}_{4}=(1,4,1,2,2,4,2,4,1,1)$  & 1.03 $\pm$ 0.13 & 0.48$\pm$ 0.05 & 0.51$\pm$ 0.05 & -0.32$\pm$ 0.04\\
      $\vec{v}_{5}=(1,4,1,3,2,3,1,4,1,1)$  & 22.50 $\pm$ 1.65 & 0.31$\pm$ 0.05& 0.68$\pm$0.05 & -0.39$\pm$0.09\\
      $\vec{v}_{6}=(1,4,1,3,2,3,1,4,2,1)$  & 18.93 $\pm$ 1.40 & 0.21$\pm$ 0.05 & 0.79$\pm$0.05 & -0.20$\pm$0.08\\
      $\vec{v}_{7}=(1,4,1,3,2,4,1,4,1,1)$  & 9.26 $\pm$ 0.71 & 0.39 $\pm$0.04 & 0.61$\pm$0.04 & -0.25$\pm$0.07\\
      $\vec{v}_{8}=(1,4,1,3,2,4,1,4,2,1)$  & 4.71 $\pm$ 0.39 &0.29 $\pm$ 0.03& 0.70$\pm$0.04 & -0.07$\pm$0.04\\
      $\vec{v}_{9}=(1,4,1,3,2,4,2,4,1,1)$  & 7.25 $\pm$ 0.57 & 0.42$\pm$ 0.04& 0.57$\pm$ 0.04 & -0.18$\pm$ 0.07\\
      $\vec{v}_{10}=(1,4,1,3,2,4,2,4,2,1)$  & 1.75 $\pm$ 0.17 & 0.35 $\pm$ 0.03& 0.64$\pm$0.05 & -0.03$\pm$0.02\\
      $\vec{v}_{11}=(1,4,1,3,3,3,1,4,2,1)$  & 1.57 $\pm$ 0.17 & 0.15 $\pm$ 0.03& 0.84$\pm$0.07 & -0.10$\pm$0.06\\
      $\vec{v}_{12}=(1,4,1,4,3,3,1,4,2,1)$ &  6.27 $\pm$ 0.50 &0.10 $\pm$ 0.04& 0.89$\pm$0.05 & -0.06$\pm$0.05\\
      $\vec{v}_{13}=(1,4,2,3,2,4,2,4,1,1)$ &  6.07 $\pm$ 0.48 & 0.46$\pm$ 0.03& 0.54$\pm$0.04 & -0.10$\pm$0.06\\
      $\vec{v}_{14}=(1,4,2,3,2,4,2,4,1,2)$ &  1.10 $\pm$ 0.12 &0.48 $\pm$ 0.05& 0.51$\pm$0.05 & -0.04$\pm$0.03\\
      $\vec{v}_{15}=(1,4,2,3,2,4,2,4,2,1)$ &  0.19 $\pm$ 0.04 &0.36 $\pm$ 0.08& 0.58 $\pm$0.13 & -0.009 $\pm$0.007\\\hline   
    \end{tabular}
\end{table}

By analyzing Table 5, we see that the variability in the estimation of $SAI(t)$ does not change the prioritization of the scenarios. The variability in the estimation of the interaction and Shapley indices is also low, which shows that $\vec{I}^{c}_{t} $ is a suitable measure to differentiate the scenarios.

\begin{figure}[htb!]
    \centering
    \includegraphics[width=0.7\textwidth]{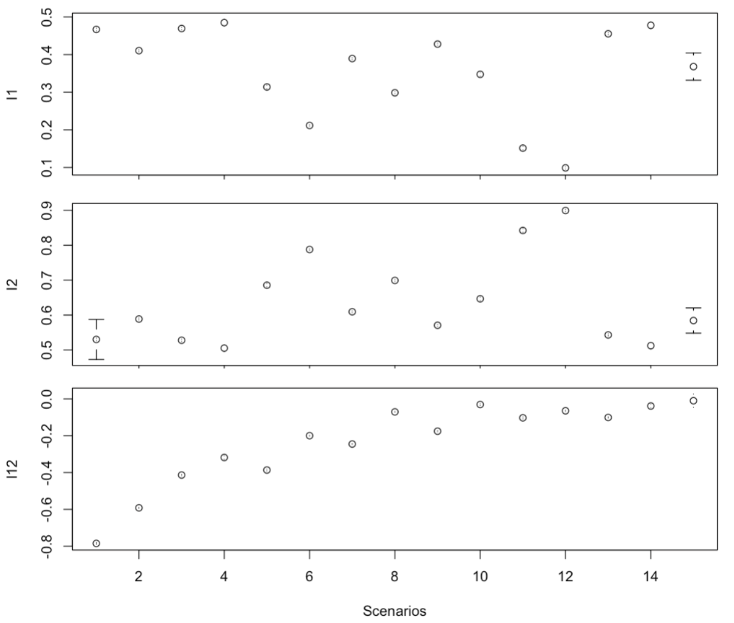}
    \caption{Confidence interval of $\vec{I}^{c}_{t}$ of each scenario.}
    \label{fig:interval_capacities}
\end{figure}

\begin{figure}[htb!]
    \centering
    \includegraphics[width=0.8\textwidth]{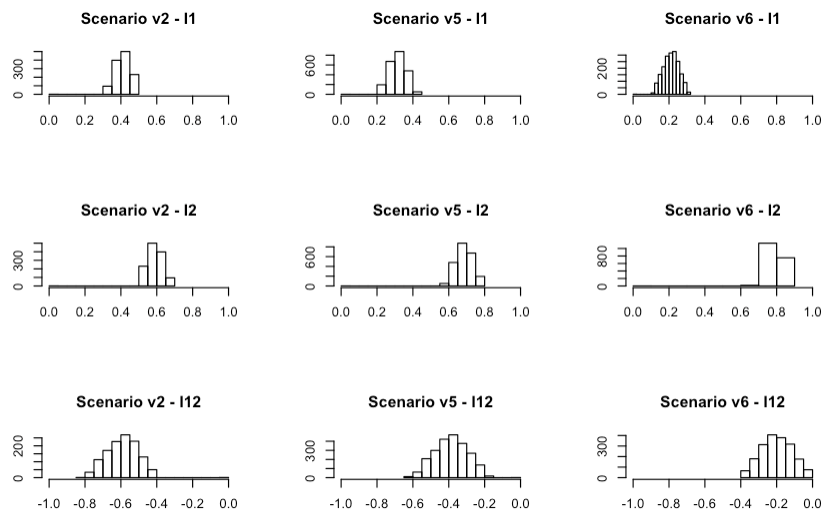}
    \caption{Distribution of the elicit capacities ($\vec{I}^{c}_{t}$) of scenarios $\vec{v}_{2}$, $\vec{v}_{5}$ and $\vec{v}_{6}$.}
    \label{fig:hist}
\end{figure}

The low variability of the $\vec{I}^{c}_{t} $ estimation also indicates that this is a stable measure for estimating capacities for each scenario. The confidence intervals of $\vec{I}^{c}_{t}$ are shown in Figure \ref{fig:interval_capacities}. On the x-axis we have the scenarios and on the y-axis the estimations of $\vec{I}^{c}_{t}$. One can also see that the variability is higher for scenarios with low SAI, as occurs for scenario $\vec{v}_{15}$. In addition, we also analyzed the distribution of the elicit capacities ($\vec{I}^{c}_{t}$) of the three scenarios with the highest SAI (Figure \ref{fig:hist}). We can see that the histograms are centralized and close to a normal distribution, indicating that using a measure such as $\vec{I}^{c}_{t}$, which is computed based on the mean, is appropriate.

\section{Conclusion}
    
    In this paper, a new capacity identification method was proposed for sorting problems based on stochastic inverse analysis. The scenario concept was introduced for SMAA sorting methods and new descriptive measures were proposed. Experiments with synthetic data were carried out in order to demonstrate the effectiveness of the proposed method. The presented experiment comprised the analysis of the proposed method considering uncertainty only in the capacities, but other experiments considering uncertain decision matrix and limiting profiles shall also be conducted in the future. 
    
    
\section*{Acknowledgments}
    \noindent This study was funded by grant \#2018/23447-4, São Paulo Research Foundation (FAPESP). L.T. Duarte thanks the National Council for Scientic and Technological Development (CNPq, grant number 311357/2017-2) for funding his research.
   
\appendix

\renewcommand{\bibname}{Bibliografia} 
\addcontentsline{toc}{chapter}{Bibliografia}
\bibliography{Ref}    

\end{document}